\documentclass[11pt]{article}

\usepackage{acl}

\usepackage{times}
\usepackage{latexsym}

\usepackage[T1]{fontenc}
\usepackage[utf8]{inputenc}

\usepackage{microtype}

\usepackage{inconsolata}

\usepackage{graphicx}
\usepackage{amsmath} % Added for math environments and commands
\usepackage{amssymb} % Added for math symbols like \mathbb
\usepackage{booktabs}
\usepackage{xcolor}
\usepackage{multirow} % Added for multirow command
\usepackage{algorithm}
\usepackage{algorithmic}
\usepackage{url}

\title{Persona-aware and Explainable Bikeability Assessment:\\A Vision-Language Model Approach}

\author{Yilong Dai\textsuperscript{1}*, Ziyi Wang\textsuperscript{2}*, Chenguang Wang\textsuperscript{3}, \\ \textbf{Kexin Zhou\textsuperscript{4}, Yiheng Qian\textsuperscript{4}, Susu Xu\textsuperscript{5}, Xiang Yan\textsuperscript{4}} \\ \textsuperscript{1}University of Alabama, \textsuperscript{2}University of Maryland, College Park, \\ \textsuperscript{3}Stony Brook University, \textsuperscript{4}University of Florida, \textsuperscript{5}Johns Hopkins University\\ \texttt{ydai17@crimson.ua.edu, zoewang@umd.edu, } \\ \texttt{chenguangwang@stonybrook.edu, } \texttt{zhoukexin@ufl.edu, } \texttt{yihengqian@ufl.edu, } \\ \texttt{susuxu@jhu.edu, xiangyan@ufl.edu}}

\begin{document}

\maketitle

\begin{abstract}
Bikeability assessment is essential for advancing sustainable urban transportation and creating cyclist-friendly cities, and it requires incorporating users’ perceptions of safety and comfort. Yet existing perception-based bikeability assessment approaches face key limitations in capturing the complexity of road environments and adequately accounting for heterogeneity in subjective user perceptions. This paper proposes a persona-aware Vision-Language Model framework for bikeability assessment with three novel contributions: (i) theory-grounded persona conditioning based on established cyclist typology that generates persona-specific explanations via chain-of-thought reasoning; (ii) multi-granularity supervised fine-tuning that combines scarce expert-annotated reasoning with abundant user ratings for joint prediction and explainable assessment; and (iii) AI-enabled data augmentation that creates controlled paired data to isolate infrastructure variable impacts. To test and validate this framework, we developed a panoramic image-based crowdsourcing system and collected 12,400 persona-conditioned assessments from 427 cyclists. Experiment results show that the proposed framework offers competitive bikeability rating prediction while uniquely enabling explainable factor attribution. 

%We further refine explanation quality through preference optimization. 

%However, existing work fails to fully address two fundamental challenges: (i) systematically measuring complex road environments with multidimensional attributes and interacting factors that affect bikeability, and (ii) accounting for subjective differences in how cyclists with varying experiences, preferences, and attitudes perceive identical infrastructure. 
\end{abstract}

\noindent\textbf{Code \& Data:} \url{https://github.com/Dyloong1/Bikeability.git}

\begin{figure}[ht!]
    \centering
    \includegraphics[width=1\linewidth]{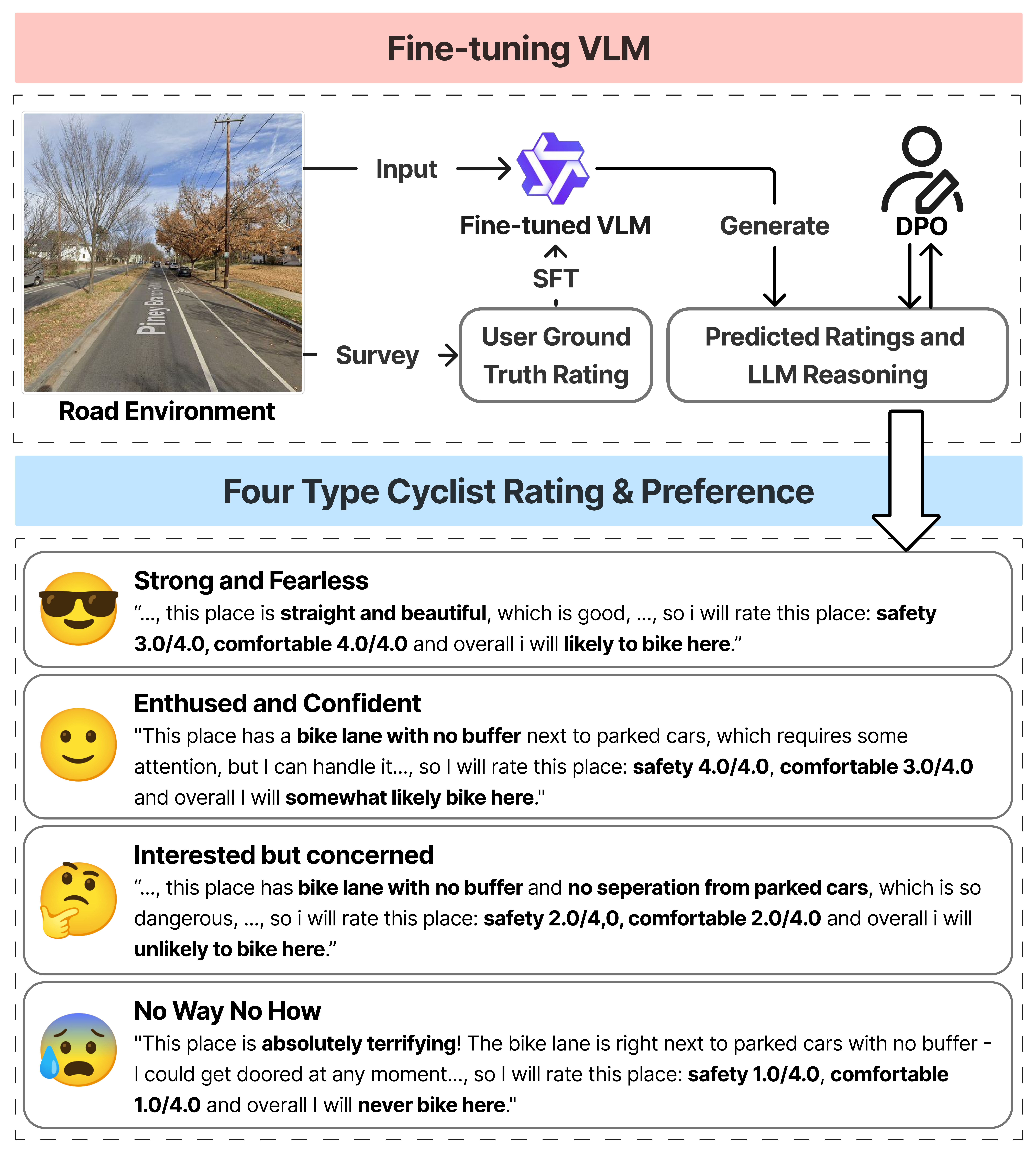}
    \caption{Overview of our persona-aware explainable bikeability assessment framework.}
    \label{fig:framework}
\end{figure}

\section{Introduction}

Bikeability assessment, which refers to the systematic evaluation of roadway infrastructure to support cycling, is crucial for advancing sustainable urban transportation and creating cyclist-friendly cities. While traditional approaches focus on objective infrastructure measures such as traffic capacity and geometric design~\cite{dowling2008multimodal,manual2000highway}, growing evidence shows that people's willingness to cycle at a location depends more by how they perceive the environment than by these objectively measured features~\cite{guo2021role,ma2014objective}. Meanwhile, people’s perceived safety at a given location can predict underlying safety risks, often identifying hazardous conditions before incidents occur~\cite{prajapati2013study,von2022crash}. Hence, there has been a growing call to incorporate users' perceptions of safety and comfort into bikeability assessment~\cite{kellstedt2021scoping,gossling2022subjectively,rodriguez2022towards,griswold2018behavioral}. However, existing perception-based assessment approaches are limited in two key aspects: (i) comprehensively measuring complex road environments with multidimensional attributes and interacting factors that affect bikeability; (ii) properly accounting for cyclist heterogeneity as cyclists with varying experiences, preferences, and attitudes may perceive identical environment differently.

%In this work, we focus on bikeability assessment as a representative application, given its significant influence on cycling decisions~\cite{winters2016} and sustainable urban development, and propose a framework that jointly models objective environmental factors and subjective user perceptions.

Recent advances in Vision-Language Models (VLMs) offer new opportunities for perception-based bikeability assessments by facilitating the generation of high-fidelity user safety and comfort perception data at scale. VLMs have demonstrated strong capabilities in capturing and mimicking human perceptions and cognitive processes, enabling various applications such as environmental assessment~\cite{ito2021assessing}, street design visualization~\cite{wang2025imagegenerationinfrastructuredesign}, and human perception modeling~\cite{danish2025citizen}. These models can process multi-modal inputs combining visual and structured data, and generate natural language outputs that facilitate interpretability. Meanwhile, persona-based methods have shown promise in capturing diverse user behaviors through LLM conditioning~\cite{wang2024largelanguagemodelsurban,chen2024personapersonalizationsurveyroleplaying}. %These advances motivate exploring VLM-based approaches for perception-based infrastructure assessment.

However, applying VLMs for perception-based bikeability assessment face several key technical challenges. First, VLMs can exhibit issues such as hallucinations and unstable predictions, which limit their reliability and generalizability. Second, 
Effective training of VLMs that balances high inference accuracy with strong interpretability necessitates large-scale data on bikeability ratings from diverse users, alongside high-quality “reasoning chain” data capturing the rationale for each rating. Yet acquiring both types of data at scale is costly and often infeasible. Third, real-world environmental factors are highly correlated (e.g., roads with bike lanes often have better greenery), which can limit a VLM’s reasoning capabilities if calibration relies solely on observational data, making it difficult to isolate the perceptual impact of individual infrastructure variables. 
%We propose a persona-aware bikeability assessment framework that addresses both the complexity of road environment measurement and the heterogeneity of subjective user perceptions. 

This paper proposes a persona-aware bikeability assessment framework (Figure~\ref{fig:framework}) that addresses these three challenges with the following novel approaches: (i) a theory-grounded persona conditioning approach that first classifies cyclists based on established typology~\cite{kim2023finite,dill2013four} and generates persona-specific explanations through chain-of-thought reasoning; (ii) a multi-granularity supervised fine-tuning strategy that combines data of varying annotation depth to jointly achieve rating prediction, factor identification, and interpretable reasoning generation; and (iii) an AI-enabled data augmentation method that elicits distinct user responses under various bicycle facility design scenarios, hence enabling the model to more effectively isolate the perceptual impact of individual infrastructure variables on bikeability ratings. To test and validate this framework, we developed a panoramic image-based crowdsourcing survey system to collect hundreds of cyclists' bikeability ratings for multiple road segments across diverse road environments in Washington DC.

%To support this framework, we designed and deployed a panoramic image-based crowdsourcing survey system covering diverse road environments in Washington DC. Through this system, we collected users' bikeability ratings for road segments, the factors influencing their assessments, and their infrastructure preferences for persona classification.

\section{Related Work}

\subsection{Bikeability Assessment}
Bikeability assessment requires evaluation of multidimensional attributes of the cycling environment while accounting for subjective user perceptions~\cite{kellstedt2021scoping,gossling2022subjectively}. Early approaches include the Level of Traffic Stress (LTS) framework~\cite{dill2016revisiting,mekuria2012low}, though it oversimplifies user heterogeneity~\cite{damant2014s}. Traditional statistical models prioritize causal relationships over prediction accuracy~\cite{breiman2001statistical,shmueli2010explain,zhao2020prediction}. Recent data-driven methods leverage diverse sources: Paranga and Oda~\cite{PARANGA20251452} applied PCA/CFA on survey indicators, Zhang et al.~\cite{zhang2025hybrid} combined COPRAS with machine learning, Ito and Biljecki~\cite{ito2021assessing} trained LightGBM on CV features, Zeng et al.~\cite{zeng2024measuring} used Random Forest, and AutoLTS~\cite{lin2024autolts} employed contrastive learning. However, these approaches either treat cyclists as homogeneous or rely on expert-predetermined weights that may not reflect actual user priorities, and vision-based methods typically require multi-stage pipelines limiting generalizability.

\subsection{Vision-Language Models for Urban Analysis}
Multi-modal approaches integrating street-view imagery, remote sensing, and geospatial data have enabled diverse urban analyses~\cite{gebru2017using,dai2025using,zhao2023sensing,danish2025citizen,suel2021multimodal,albert2017using,cao2020deep}. Open-source VLMs such as Qwen-VL~\cite{bai2025qwen3vl} and LLaVA~\cite{liu2024improved} now provide strong visual understanding with parameter-efficient fine-tuning capabilities~\cite{hu2022lora}, enabling domain adaptation for specialized tasks. These models advance urban analysis through CoT reasoning~\cite{wei2022chain,kong2024controllable} and multi-granularity instruction tuning~\cite{liu2024improved,yang2025curriculum,wang2022self,sanh2021multitask,chung2024scaling}, showing promise in navigation~\cite{wang2025navrag,wu2025goviggoalconditionedvisualnavigation}, infrastructure assessment~\cite{ito2021assessing}, street design~\cite{wang2025imagegenerationinfrastructuredesign}, and decision-making~\cite{chen2025perceptionsdecisionswildfireevacuation}. Meanwhile, recent advances in image generation models~\cite{yang2025textimagegenerationediting} have enabled high-fidelity synthetic data creation, offering new possibilities for controlled data augmentation in domains where paired observations are scarce. However, domain applications require interpretability for stakeholder validation~\cite{arrieta2020explainable,javed2023survey,dong2025large}, and existing approaches focus on synthetic or web-scale data rather than domain-specific settings where expert annotations are scarce. For such subjective perception tasks, aligning model outputs with human preferences is critical, as ground truth reflects individual judgments rather than objective labels. Preference optimization methods such as RLHF~\cite{ouyang2022training} and DPO~\cite{rafailov2023direct,bai2022constitutional,meng2024simpo,xiao2025dposurvey} provide effective mechanisms for this alignment.

\subsection{Persona-aware Modeling}
User heterogeneity is fundamental to perception-based assessment, as expert-derived standards often deviate from actual user experiences~\cite{gossling2022subjectively}. The ``Four Types of Cyclists'' typology~\cite{dill2013four,dill2016revisiting} confirms significant differences in infrastructure needs across cyclist types. Persona-conditioning enhances LLM behavioral simulation~\cite{wang2024largelanguagemodelsurban,li2024realtraveldiarygeneration,chen2024personapersonalizationsurveyroleplaying,sun2025personadbefficientlargelanguage}, yet existing methods rely on predefined categories or aggregated data~\cite{chuang2024demographicsaligningroleplayingllmbased,park2024generativeagentsimulations1000}, lacking mechanisms to model differentiated visual interpretations of streetscapes~\cite{ito2021assessing}.

\begin{figure*}[t!]
\centering
\includegraphics[width=1\linewidth]{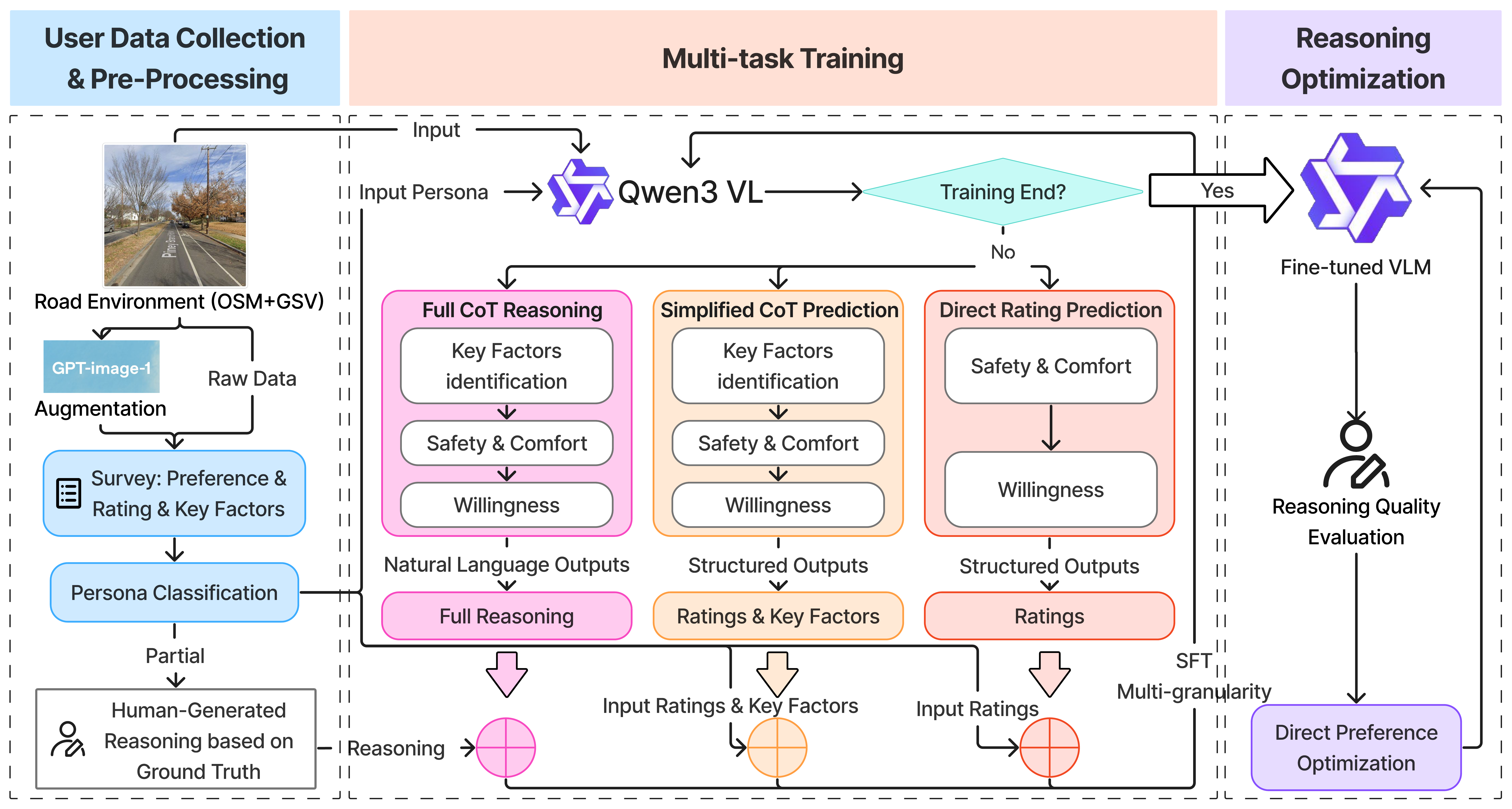}
\caption{Three-stage model architecture: (1) Multi-source data collection with crowdsourced survey and AI-based image augmentation, (2) Multi-granularity supervised fine-tuning with three data types (Type 1: full reasoning, Type 2: factor-rating pairs, Type 3: rating-only), and (3) Preference-based reasoning refinement with DPO.}

\label{fig:Architecture}
\end{figure*}

\section{Method}

We propose a persona-aware bikeability assessment framework that addresses both the complexity of objective road environment assessment and the heterogeneity of subjective user perceptions. Figure~\ref{fig:Architecture} illustrates our pipeline, which consists of three stages: (1) Multi-source Data Collection: We gather persona-specific bikeability ratings and influencing factors through crowdsourcing surveys covering diverse Washington DC road environments, with AI-based image augmentation for controlled infrastructure variations. (2) Multi-Granularity Supervised Fine-tuning: We create three types of training data (expert-annotated full reasoning chains, user-provided structured factor-rating pairs, and direct ratings) and jointly train on all three with fixed sampling ratios, enabling the model to learn interpretable reasoning while maintaining robust prediction. (3) Preference-based Reasoning Refinement: We refine reasoning outputs through Direct Preference Optimization (DPO) based on human feedback.

\subsection{Data Collection and Preprocessing}

\subsubsection{Study Design and Sampling}
We randomly sampled 200 road segments from Washington DC's road network, prioritizing spatial dispersion to ensure diverse cycling environments. The selected segments encompass varying motor vehicle lane configurations, cycling infrastructure types (from absent to fully protected), and urban contexts. We integrated GSV panoramic API into our survey system, enabling participants to conduct detailed 360-degree observations rather than being limited to static 2D images.

\subsubsection{Image Augmentation Pipeline}
\begin{figure}[h]
    \centering
    \includegraphics[width=1\linewidth]{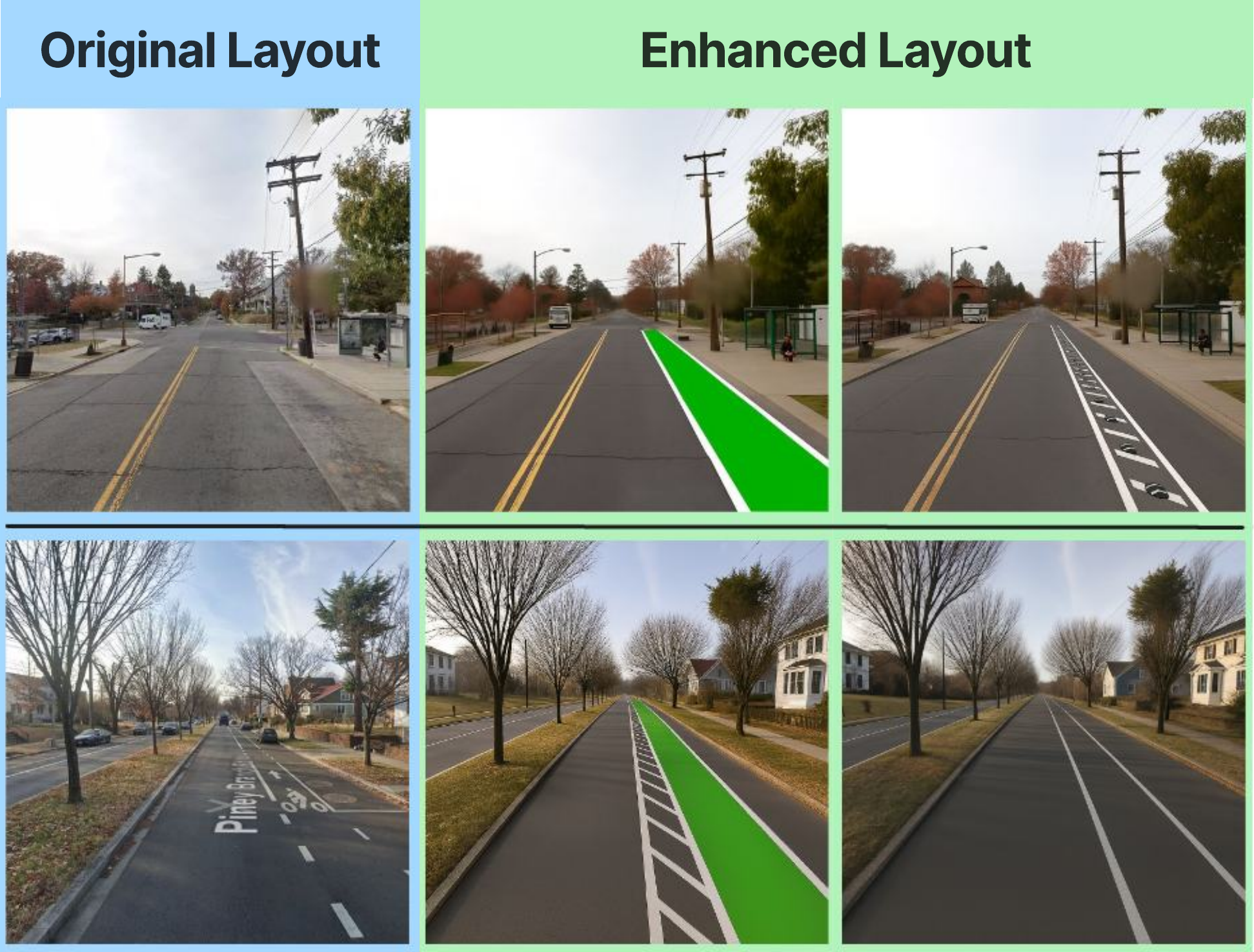}
    \caption{AI-based image augmentation: original street-view images (left) and systematically modified versions with controlled infrastructure changes (right).}
    \label{fig:exp_enhance}
\end{figure}

During our road segment sampling process, we found that natural road networks rarely provide ideal paired comparisons where segments differ in only a minimal number of infrastructure attributes, even adjacent roads typically vary across multiple dimensions simultaneously. This makes it challenging for models to learn fine-grained attribute importance from limited survey data. To address this, we developed an AI-based image editing pipeline using GPT-image-1. This pipeline systematically modifies objectively defined infrastructure variables without introducing subjective developer judgments, including: Bike Lane Presence, Lane Width (Narrow/Standard/Wide), Lane Color (Green/No Paint), Buffer Type (No Buffer/Standard/Bollards/Armadillo), and Buffer Location (Adjacent to Moving Cars/Adjacent to Parked Cars). By altering these variables while preserving all other environmental factors, the pipeline enables valid paired comparisons for isolating infrastructure impacts. The original image sample and image editing result is shown in Figure~\ref{fig:exp_enhance}.

\subsubsection{Crowdsourcing survey}
Our survey collects the following information:

\textit{Demographics}: Gender, age group, race/ethnicity, and education level to ensure diverse representation.

\textit{Cycling preferences}: Comfort ratings (5-point scale: 1=Very Uncomfortable to 5=Very Comfortable) for eight infrastructure types using reference images: (1) no bike lanes, (2) roadway shoulders, (3) off-street multi-use paths, (4) shared lanes with sharrows, (5) sidewalks, (6) striped bike lanes, (7) buffered bike lanes, and (8) protected bike lanes. These ratings enable cyclist persona identification, capturing individual risk tolerance and comfort thresholds.

\textit{Segment evaluations}: For 15 locations (20 after augmentation) randomly assigned to each respondent, we collected: (1) \textit{Safety rating} and (2) \textit{Comfort rating} on 4-point Likert scales (1=Strongly Disagree to 4=Strongly Agree); (3) \textit{Cycling willingness} on a 4-point scale (1=Never to 4=Absolutely); and (4) \textit{Influencing factors} via a multi-select tag interface with predefined options (bike lane conditions, traffic, separation, environment) and open-ended fields.

%Through this design, we balance data collection efficiency with comprehensiveness, enabling
\subsubsection{Persona Classification}
\label{sec:persona_classification}

Our survey collected responses from 427 participants. Each participant first provided comfort ratings (1-5 scale) for eight infrastructure types, then rated a minimum of 20 street-view images for bikeability assessment, resulting in 12,400 total persona-conditioned assessments (mean: 29.0 images per participant).

\begin{figure}[h]
    \centering
    \includegraphics[width=1\linewidth]{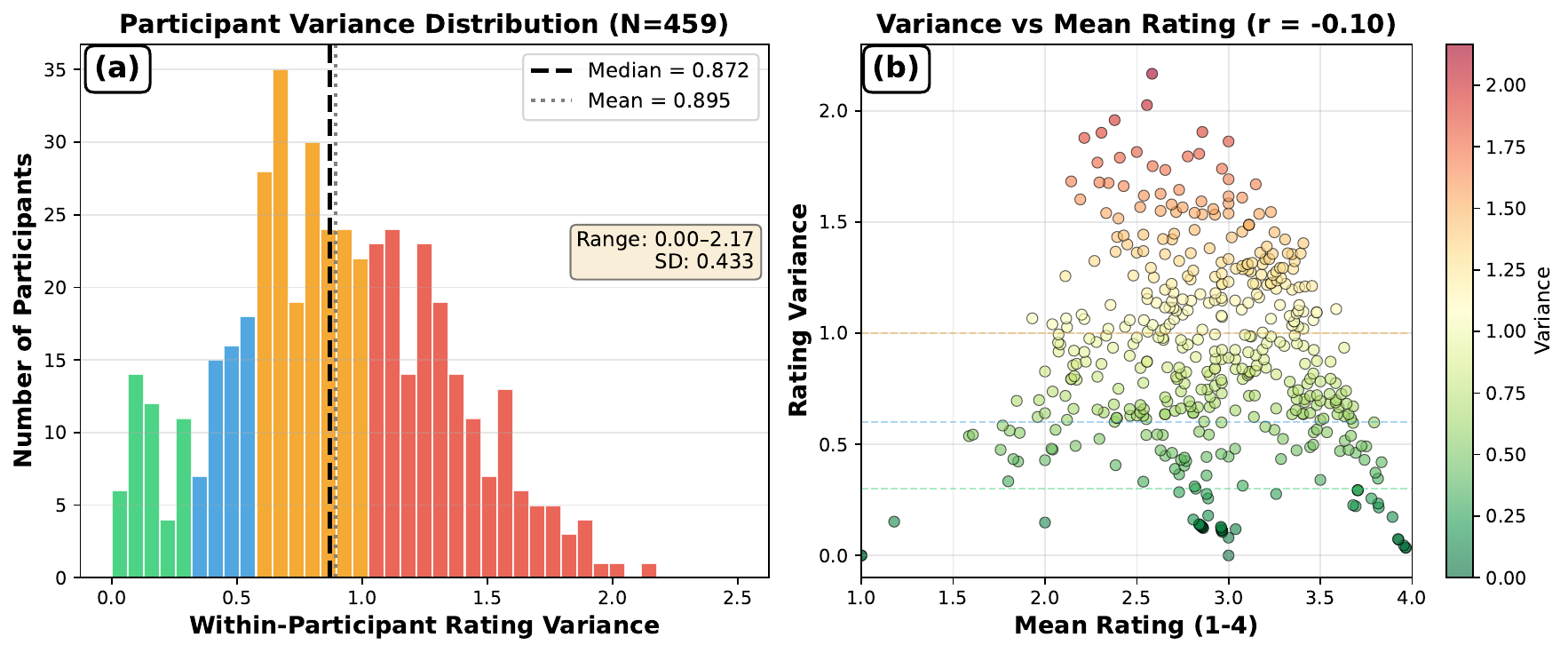}
    \caption{Within-participant rating variance distribution across 427 participants. (a) Histogram shows substantial heterogeneity (median = 0.872, range: 0--2.17). (b) Scatter plot confirms variance is orthogonal to mean rating (r = $-$0.10).}
    \label{fig:participant_variance}
\end{figure}

To validate the necessity of persona-aware modeling, we analyzed within-participant rating variance. Figure~\ref{fig:participant_variance} reveals substantial inter-participant heterogeneity: variance in street segment ratings ranges from 0 to 2.17 (median = 0.872), and is orthogonal to mean rating levels (r = $-$0.10), confirming genuine individual differences rather than scale-use artifacts.

We classified participants into four cyclist personas following the ``Four Types of Cyclists'' typology~\cite{dill2016revisiting}. We computed mean comfort ratings under three protection levels: low (no bike lanes, shared lanes), medium (striped lanes, shoulders, off-street paths), and high (buffered/protected lanes), plus the gradient from low to high protection. Clustering on these indicators reveals four groups: participants with low overall comfort and minimal gradient emerge as \textit{No Way No How} (NWNH); those with high comfort and flat gradients cluster as \textit{Strong \& Fearless} (S\&F); individuals with steep gradients form the \textit{Interested but Concerned} (IbC) group; while those with moderate comfort and gradients constitute \textit{Enthused \& Confident} (E\&C). Table~\ref{tab:persona_characteristics} presents the distribution.

\begin{table}[h]
\centering
\caption{Persona distribution and infrastructure preference characteristics (N=427). Mean and Grad (gradient) are computed from comfort ratings (1-5 scale) for eight infrastructure types.}
\label{tab:persona_characteristics}
\begin{tabular}{lccc}
\toprule
\textbf{Persona} & \textbf{Pct.} & \textbf{Mean} & \textbf{Grad.} \\
\midrule
IbC & 59.3\% & 2.97 & 2.73 \\
E\&C & 27.6\% & 3.45 & 1.57 \\
S\&F & 8.2\% & 3.80 & 0.13 \\
NWNH & 4.9\% & 2.02 & 0.50 \\
\bottomrule
\end{tabular}
\end{table}

\subsection{Task Definition}

We formulate persona-aware bikeability assessment as a conditional generation task where VLM generates subjective evaluations based on cyclist personas and multi-modal inputs. By combining visual street-level observations with structured infrastructure attributes as input, we ensure comprehensive coverage while avoiding human selection bias in feature engineering.

Formally, given a cyclist persona $P$, the model generates a subjective evaluation text $X = \{x_1, x_2, ..., x_S\}$ consisting of $S$ words that describe the bikeability assessment from the perspective of that persona type. The model takes as input:
\begin{itemize}
    \item A cyclist persona $P$ from four types: S\&F, E\&C, IbC, or NWNH
    \item $O$: A street-view image capturing the first-person cyclist perspective, providing rich visual context
    \item $A = \{a_1, a_2, ..., a_I\}$: A set of $I$ infrastructure attributes from OpenStreetMap (e.g., road type, lane configuration), ensuring comprehensive coverage of non-visual factors
\end{itemize}

Given the ground truth user evaluation $X^*$ from our survey data, we optimize the model parameters $\theta$ to maximize the conditional probability of generating the correct evaluation:

\begin{equation}
\theta^* = \underset{\theta}{\operatorname{argmax}} \log p(X^* | P, O, A; \theta)
\end{equation}

Following established transportation theory~\cite{dill2016revisiting,winters2011motivators} and our empirical validation of systematic inter-participant heterogeneity (Section~\ref{sec:persona_classification}, Figure~\ref{fig:participant_variance}), we treat persona-specific assessment as a fundamental design requirement rather than an optional feature.

\subsubsection{Multi-format Training Data}

Since manually annotated complete Chain-of-Thought (CoT) reasoning is scarce and labor-intensive, we design three training data formats with varying supervision granularity. Our multi-granularity strategy leverages complementary strengths: Type 1 enables interpretable reasoning generation, Type 2 develops factor identification skills, and Type 3 ensures robust rating prediction. By jointly training on all three types with fixed sampling ratios (Section~\ref{sec:exp_setup}), the model learns a unified representation supporting multiple output formats while maintaining consistency across explanation depths. All three formats share identical multi-modal inputs $(P, O, A)$ but differ in their prompt instructions and supervision signals.

\textbf{Type 1: Full CoT Reasoning.}
This format trains the model to generate interpretable natural language reasoning. However, obtaining complete reasoning chains directly from crowdsourced surveys is challenging due to: (i) \textit{cognitive burden} leading to lower completion rates~\cite{krosnick1991response,deutskens2004response}; (ii) \textit{expression heterogeneity} creating noise that complicates training; and (iii) \textit{implicit reasoning} where perceptual judgments occur subconsciously~\cite{nisbett1977telling,wilson2002strangers}.

Given these constraints, we construct full reasoning chains through expert annotation by transportation researchers. Critically, this differs from traditional expert-driven approaches: rather than predefining which factors matter, experts receive user-provided factors and ratings from survey responses and synthesize them into coherent reasoning chains. This data-driven approach ensures reasoning patterns emerge from actual user preferences. The model receives:
\begin{equation}
[P, O, A; \text{prompt}_{\text{reason}}]
\end{equation}
where $\text{prompt}_{\text{reason}}$ instructs the model to: (i) analyze environmental conditions from the image, (ii) identify key factors influencing the cyclist's perception, (iii) predict safety and comfort ratings, and (iv) synthesize an overall willingness assessment. The supervision signal is:
\begin{equation}
X^{*}_{\text{reason}} = \text{Annotate}(T^*, R^*)
\end{equation}
where transportation researchers compose reasoning chains connecting the ground truth factors $T^*$ and ratings $R^*$ using natural language. While large teacher models (e.g., GPT-4) could automate annotation, such approaches introduce hallucination risks and inconsistent patterns~\cite{bang2023multitask,xu2024critical}. Our expert annotation ensures high-quality supervision that faithfully represents participant perspectives.

\textbf{Type 2: Simplified CoT Prediction.}
Unlike Type 1, this format lacks manually completed reasoning narratives and is derived directly from survey responses. The model receives:
\begin{equation}
[P, O, A; \text{prompt}_{\text{struct}}]
\end{equation}
where $\text{prompt}_{\text{struct}}$ instructs the model to perform a simplified CoT process that omits the objective condition analysis step: (i) identifying key factors that most influence the cyclist's perception, (ii) predicting safety and comfort ratings, and (iii) providing an overall willingness-to-use assessment. The supervision signals consist of paired annotations from participants:
\begin{equation}
(T^*, R^*)
\end{equation}
where $T^* = \{t_1, t_2, ..., t_n\}$ represents the set of influencing factors identified by participants , and $R^*$ contains their corresponding three-dimensional ratings. This format bridges perception and evaluation through explicit factor identification without requiring full explanatory narratives.

\textbf{Type 3: Direct Rating Prediction.}
This format further simplifies the task by removing the factor identification step, utilizing all available survey ratings for maximum data efficiency. The model receives:
\begin{equation}
[P, O, A; \text{prompt}_{\text{rating}}]
\end{equation}
where $\text{prompt}_{\text{rating}}$ requests direct rating prediction: the model predicts safety and comfort ratings, followed by an overall willingness-to-use assessment, without requiring intermediate factor identification or reasoning. The supervision signal consists of the three-dimensional rating vector:
\begin{equation}
R^* = \{r_{\text{safety}}, r_{\text{comfort}}, r_{\text{willingness}}\}
\end{equation}
This format represents basic stimulus-response mapping, providing the most direct supervision signal and ensuring robust end-to-end prediction capabilities.

\subsubsection{Model Training}

To efficiently adapt the model given limited survey data and prevent overfitting, we employ parameter-efficient fine-tuning using LoRA \cite{hu2022lora} on the pretrained Qwen3-VL-8B-Instruct base model.

\textbf{Multi-Granularity Instruction Tuning.}
Since manually annotated complete CoT reasoning (Type 1) is scarce and labor-intensive, we adopt a multi-granularity instruction tuning approach following LLaVA-1.5~\cite{liu2024improved}. Our training data exhibits natural granularity variation: while all 12,400 survey responses provide ratings (Type 3) and user-identified factors (Type 2), only approximately 2,000 samples contain expert-annotated reasoning chains (Type 1). This data constraint necessitates multi-granularity training: relying solely on Type 1 would provide insufficient samples for learning robust visual-linguistic representations; conversely, using only Type 3 would sacrifice all interpretability. Our multi-granularity strategy leverages complementary strengths: Type 1 enables interpretable reasoning generation, Type 2 develops factor identification skills, and Type 3 ensures robust rating prediction. By jointly training on all three types with fixed sampling ratios (Section~\ref{sec:exp_setup}), the model learns a unified representation supporting multiple output formats while maintaining consistency across explanation depths.

\textbf{Direct Preference Optimization.}
Following supervised fine-tuning, we apply Direct Preference Optimization (DPO)~\cite{rafailov2023direct} to Type 1 outputs to refine reasoning quality. To collect preference data, we sample instances from the training set and use the SFT model to generate two Type 1 reasoning explanations per instance with different sampling temperatures. Three transportation domain experts independently evaluate both explanations, selecting the preferred one based on: (i) \textit{Factual Accuracy}, (ii) \textit{Logical Coherence}, and (iii) \textit{Persona Consistency}. Majority voting determines the final preference pairs. Following the DPO framework, we directly optimize the policy to increase the likelihood of preferred responses without training a separate reward model. The DPO objective uses reference model regularization to prevent deviation from the SFT initialization, preserving performance on Type 2 and Type 3 formats.

\section{Experiments}

\subsection{Experimental Setup}
\label{sec:exp_setup}

\textbf{Implementation Details.}
We conduct all experiments on one NVIDIA A100 80GB GPU. For supervised fine-tuning, we employ LoRA with rank $r=32$ and scaling factor $\alpha=64$, targeting the query, key, value, and output projection matrices (\texttt{q\_proj}, \texttt{k\_proj}, \texttt{v\_proj}, \texttt{o\_proj}). We use AdamW optimizer with an initial learning rate of $2 \times 10^{-4}$ and train for 5 epochs with cosine annealing learning rate schedule. The learning rate follows a decay schedule: gentle decay ($0.8\times$ per epoch) for the first 3 epochs, then rapid decay ($0.5\times$ per epoch) for the remaining epochs. We train with batch size 4 and gradient accumulation over 4 steps for an effective batch size of 16. We apply gradient checkpointing to manage memory usage and use 10\% warmup steps. Mixed precision training (FP16) is employed for efficiency.

\textbf{Multi-Granularity Sampling Ratios.}
We sample training data with fixed ratios across all epochs: 15\% Type 1 (full reasoning), 40\% Type 2 (structured factor-rating pairs), and 45\% Type 3 (rating-only). This distribution is consistent with LLaVA-1.5's finding that detailed reasoning samples comprising ~15\% yields optimal performance~\cite{liu2024improved}. The fixed-ratio design ensures: (i) efficient utilization of scarce expert annotations without over-sampling (15\% yields $\sim$1,860 Type 1 samples per epoch, approaching our full 2,000-sample budget), (ii) balanced multi-task learning across all supervision levels, and (iii) training stability by avoiding catastrophic forgetting~\cite{kirkpatrick2017overcoming}. We enforce strict rating constraints (1-4 scale) through explicit prompt instructions and post-processing corrections.

\textbf{DPO Training Details.}
For preference data collection, we randomly sample 500 instances and generate two explanations per instance with sampling temperatures 0.7 and 1.0, resulting in 500 preference pairs after expert majority voting. For DPO training, we use regularization coefficient $\beta=0.1$, learning rate $5 \times 10^{-6}$, batch size 8, and train for 3 epochs.

\textbf{Baselines.}
We compare against three baselines: (1) \textit{GPT-4o Zero-shot}: prompting GPT-4o with persona descriptions and multi-modal inputs without task-specific training; (2) \textit{Kmeans-SMOTE RF}~\cite{zeng2024measuring}: To ensure fair comparison, we replicate Zeng et al.'s preprocessing-based approach by employing YOLOv8 for comprehensive detection of observable road infrastructure elements (bike lanes, buffers, traffic signals, street furniture, greenery, etc.), combining these computer vision features with OpenStreetMap attributes (motor vehicle lane count, speed limits, bike lane types) and image-derived latent representations (encoded via a pretrained ResNet-50). These multi-source features are then fed into a Random Forest classifier following their Kmeans-SMOTE balancing strategy, extended with our LLM-normalized factor tag pool for factor identification; (3) \textit{Kmeans-SMOTE RF (Rating-only)}: same preprocessing and feature extraction pipeline but trained only for rating prediction, isolating the contribution of joint tag-rating learning.

\textbf{Evaluation Metrics.}
For rating prediction, we report metrics averaged across all three dimensions (safety, comfort, willingness-to-use): Mean Absolute Error (MAE), Exact Match Rate (EM), Within-One Accuracy (W1, percentage within $\pm$1), and Pearson correlation. For factor identification, we compute Semantic Precision, Recall, and F1-score using sentence embeddings (all-MiniLM-L6-v2) with greedy matching at threshold 0.7. For reasoning quality, we use GPT-4o as an automatic judge~\cite{zheng2023judging} to assess: (1) \textit{Factual Accuracy}, correct description of infrastructure features; (2) \textit{Logical Coherence}, clarity and logical flow; (3) \textit{Persona Consistency}, alignment with persona-specific concerns.

\subsection{Main Results}

\begin{table*}[!ht]
\centering
\small
\caption{Main experimental results comparing rating prediction and factor identification performance. MAE: Mean Absolute Error, EM: Exact Match, W1: Within-One Accuracy, Corr: Pearson Correlation, Prec: Precision, Rec: Recall, F1: F1-score (at threshold=0.7).}
\label{tab:main_results}
\begin{tabular}{lcccc|ccc}
\toprule
\multirow{2}{*}{\textbf{Method}} & \multicolumn{4}{c|}{\textbf{Rating Prediction}} & \multicolumn{3}{c}{\textbf{Factor Identification}} \\
\cmidrule(lr){2-5} \cmidrule(lr){6-8}
& \textbf{MAE↓} & \textbf{EM↑} & \textbf{W1↑} & \textbf{Corr↑} & \textbf{Prec↑} & \textbf{Rec↑} & \textbf{F1↑} \\
\midrule
GPT-4o Zero-shot & 1.00 & 0.30 & 0.70 & 0.25 & 0.12 & 0.08 & 0.10 \\
KS-RF (Rating-only) & \textbf{0.70} & \textbf{0.45} & 0.85 & \textbf{0.50} & - & - & - \\
KS-RF & 0.80 & 0.38 & 0.82 & 0.45 & 0.33 & 0.30 & 0.31 \\
\textbf{Ours} & 0.71 & 0.41 & \textbf{0.87} & 0.48 & \textbf{0.52} & \textbf{0.46} & \textbf{0.49} \\
\bottomrule
\end{tabular}
\end{table*}

Table~\ref{tab:main_results} presents our main results. For rating prediction, our method achieves competitive performance (MAE: 0.71, EM: 0.41, Corr: 0.48) closely matching the specialized regression approach KS-RF (Rating-only), while achieving the highest within-one accuracy (W1: 0.87).

For factor identification, our approach achieves 0.52 precision, 0.46 recall, and 0.49 F1-score, substantially outperforming KS-RF (F1: 0.31) and GPT-4o zero-shot (F1: 0.10). This demonstrates that domain-specific training enables accurate identification of bikeability-relevant factors despite wording variations between generated and ground-truth factors.

\subsection{Human Preference Alignment}

\begin{table}[!ht]
\centering
\small
\caption{Human preference alignment results. Acc.: Factual Accuracy, Coh.: Logical Coherence, Cons.: Persona Consistency.}
\label{tab:dpo_results}
\begin{tabular}{lccc}
\toprule
\textbf{Method} & \textbf{Acc.↑} & \textbf{Coh.↑} & \textbf{Cons.↑} \\
\midrule
GPT-4o & 0.25 & 0.694 & 0.995 \\
Ours (SFT) & 0.58 & 0.580 & 0.920 \\
\textbf{Ours+DPO} & \textbf{0.59} & \textbf{0.610} & \textbf{0.950} \\
\midrule
\textit{Increase} & +1.7\% & +5.2\% & +3.3\% \\
\bottomrule
\end{tabular}
\end{table}

Table~\ref{tab:dpo_results} shows that DPO improves explanation quality across all dimensions. Our SFT model achieves substantially higher factual accuracy (0.58) than GPT-4o (0.25), reflecting domain-specific training, while GPT-4o exhibits stronger coherence and consistency due to its larger scale. DPO narrows this gap, with the largest improvement in coherence (+5.2\%), validating that preference optimization refines explanation style and logical flow.

\subsection{Ablation Studies}

\textbf{Multi-Granularity Data Ablation.}
We evaluate the contribution of different supervision granularities. Training with Type 3 only (rating-only) achieves MAE 0.75 and W1 0.85, but provides no factor identification or explanations. Adding Type 2 (factor-rating pairs) improves prediction to MAE 0.73 and W1 0.86, though reasoning remains absent.

Our full approach with Type 1 (CoT reasoning) at 15/40/45 ratios achieves the best accuracy (MAE: 0.71, W1: 0.87) while enabling factor identification (F1: 0.49) and interpretable reasoning. This demonstrates that CoT provides the richest supervisory signal for joint optimization. The fixed-ratio strategy ensures training stability, as aggressive epoch-wise ratio changes led to unstable convergence and catastrophic forgetting.

\section{Conclusion}

We presented a persona-aware VLM approach for bikeability assessment that addresses infrastructure evaluation complexity and user perception heterogeneity. Our results demonstrate that interpretability need not compromise accuracy: our framework achieves competitive rating prediction while uniquely enabling factor identification and interpretable reasoning, capabilities absent in existing approaches.

For researchers, our framework eliminates repetitive preprocessing pipelines when adapting to new urban contexts. For planners, persona-specific assessments reveal diverse community needs beyond the ``average'' cyclist, informing more inclusive infrastructure decisions. We hope this work inspires broader VLM adoption in transportation research, supporting human-centered infrastructure assessment that accounts for user diversity.

\section{Limitations}

Our study has several limitations. First, while the framework is transferable, the model trained on Washington DC data may require fine-tuning for cities with different infrastructure styles. Second, our four-category cyclist typology is a discrete approximation of continuous preference spectra. Third, static street-view imagery does not capture dynamic factors such as traffic volume, weather, or temporal variations. Finally, a gap remains in general language capabilities compared to larger models like GPT-4o.

\section{Acknowledgments}
We are grateful for the support from District Department of Transportation in this project. This project is supported by U.S. National Science Foundation (Award No.2425029).

\bibliography{acl2026}

\appendix

\section{Appendix}

\subsection{Recruitment And Payment}
We recruited participants through social media platforms to complete 
anonymous surveys evaluating bicycle lane quality. A total of 427 
participants completed the survey. Each survey took approximately 
5 minutes to complete. To incentivize participation, we implemented 
a lottery-based compensation system: every 100th participant received 
a \$100 gift card (4 participants total received compensation). This 
resulted in an average expected compensation of approximately \$0.93 
per participant, or approximately \$11.16 per hour based on the 
estimated completion time. Participation was voluntary and anonymous, 
with no personally identifiable information collected.
with no personally identifiable information collected.

Three graduate students were recruited from a research university to 
perform data annotation tasks. Annotators were paid \$20 per hour for 
approximately 8 hours of annotation work each, totaling \$160 per 
annotator. This compensation rate exceeds local minimum wage standards 
and is consistent with standard research assistant rates at our 
institution.
\subsection{Survey Interface}
\begin{figure}[ht!]
\centering
\includegraphics[width=\linewidth]{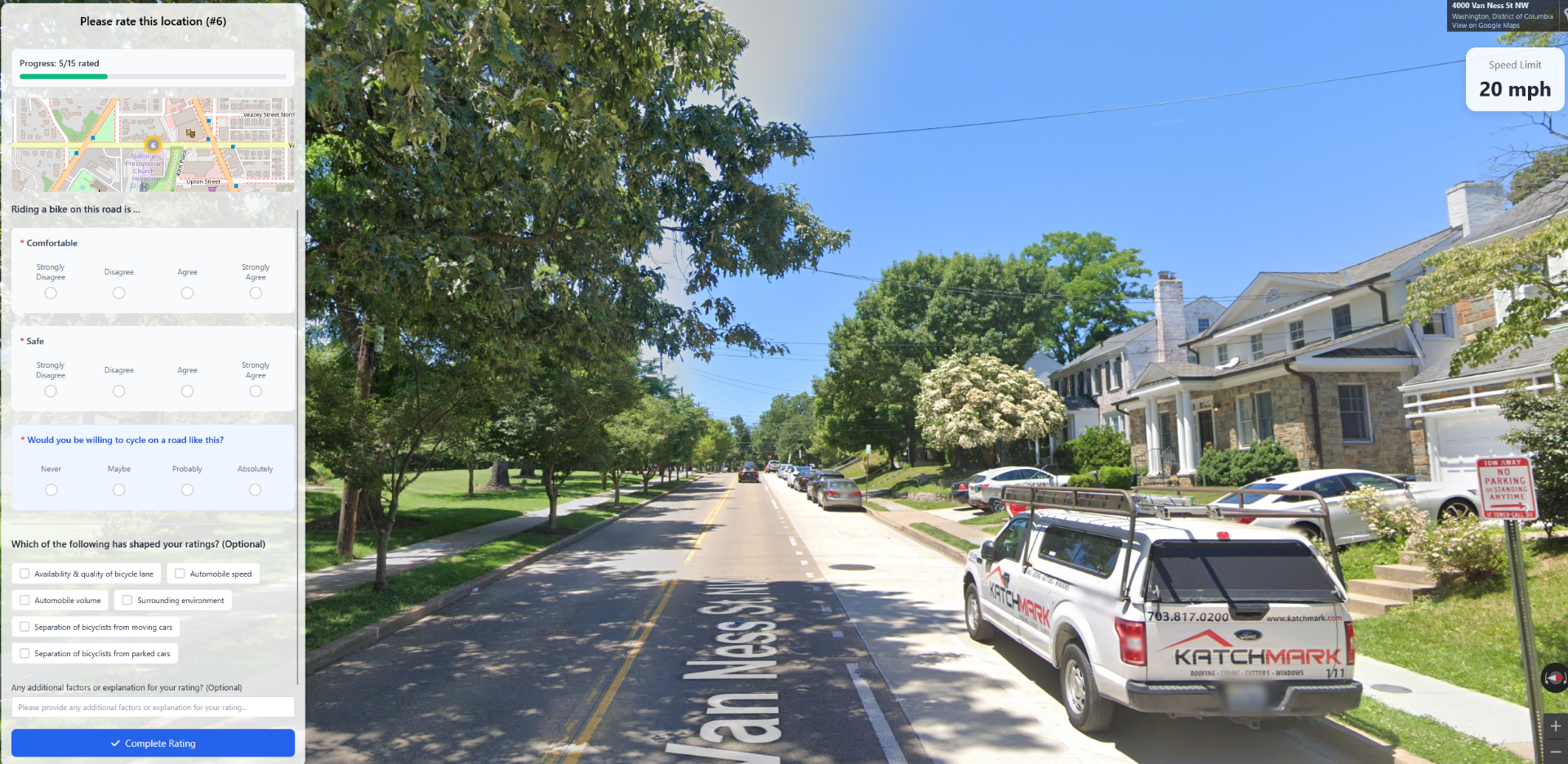}
\caption{Survey interface 1: immersive 360-degree Google Street View for bikeability assessment.}
\label{fig:gsv_view}
\end{figure}

\begin{figure}[ht!]
\centering
\includegraphics[width=\linewidth]{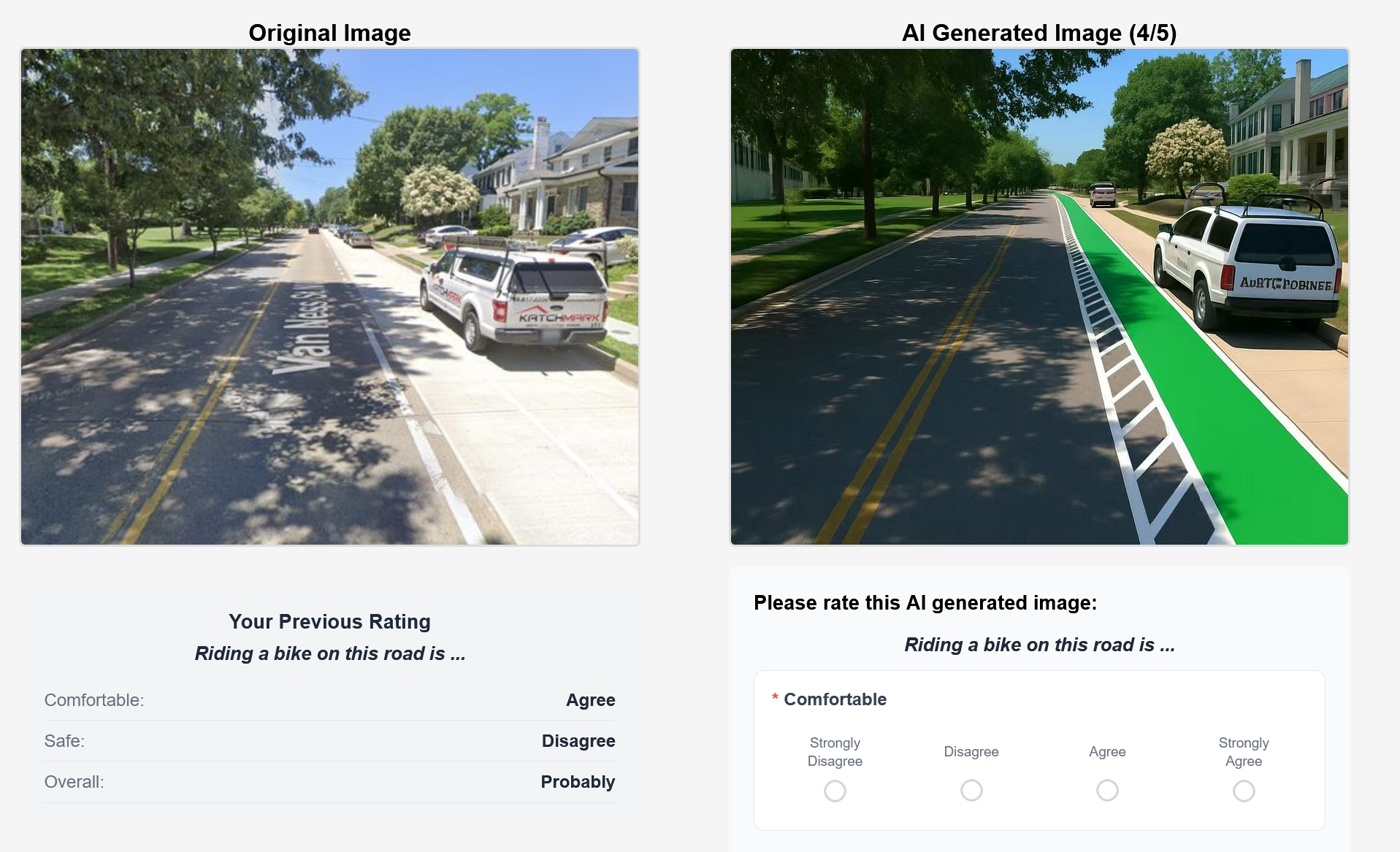}
\caption{Survey interface 2: rating for augmented image.}
\label{fig:single_var_assessment}
\end{figure}

Figures~\ref{fig:gsv_view} and~\ref{fig:single_var_assessment} show screenshots from our crowdsourcing survey platform. Figure~\ref{fig:gsv_view} displays the immersive 360-degree Google Street View interface used for bikeability assessment, allowing participants to explore road environments interactively. Figure~\ref{fig:single_var_assessment} shows the infrastructure preference assessment interface where participants rate their comfort levels for different cycling facility types.

\subsection{Prompt Templates}

We provide the prompt templates used for model inference and baseline evaluation.

\subsubsection{Type 1: Full CoT Reasoning}
\begin{verbatim}
As a {persona} cyclist ({persona_detailed_desc}),
analyze this street image for bikeability.

Provide a brief assessment covering:
- Key observations about the street
- Factors affecting your cycling experience
- Your comfort and safety evaluation

Rate the following on a scale of 1-4:
- Comfortable: How comfortable would you feel
  cycling here?
- Safe: How safe would you perceive this road?
- Overall: Your overall willingness to cycle
  on this road

End with:
STRUCTURED OUTPUT:
Factors: [list specific factors]
Ratings: comfortable: X, safe: Y, overall: Z
\end{verbatim}

\subsubsection{Type 2: Simplified CoT Prediction}
\begin{verbatim}
As a {persona} cyclist ({persona_brief_desc}),
assess this street for bikeability.

Identify the most important factors affecting
bikeability for someone with your cycling
preferences, then rate the street.

Format your response as:
Factors: [list key factors]
Ratings: comfortable: X, safe: Y, overall: Z

Use a 1-4 scale for ratings.
\end{verbatim}

\subsubsection{Type 3: Direct Rating}
\begin{verbatim}
As a {persona} cyclist ({persona_brief_desc}),
rate this street's bikeability.

Provide ratings (1-4 scale):
Ratings: comfortable: X, safe: Y, overall: Z
\end{verbatim}

\subsubsection{GPT-4o Zero-shot Baseline}
\begin{verbatim}
You are an expert in urban cycling infrastructure
assessment. Analyze the provided street view image
and assess its bikeability from the perspective
of a specific cyclist persona.

Cyclist Persona: {persona}
Persona Description: {persona_desc}
{osm_text}

Task: Perform a bikeability assessment:
1. Analyze the street environment and OSM
   attributes
2. Consider how a "{persona}" cyclist would
   perceive this environment
3. Provide ratings (1-4) for: Comfortable,
   Safe, Overall
4. Output JSON with influencing_factors and
   ratings
\end{verbatim}

\subsubsection{Persona Descriptions}
The four cyclist personas used in prompts:
\begin{verbatim}
Strong and Fearless: Comfortable with all
  infrastructure types, showing little preference
  between protected and unprotected facilities.

Enthused and Confident: Regular cyclists who
  prefer bike lanes but will ride in mixed
  traffic when necessary.

Interested but Concerned: Would cycle more if
  separated from traffic; requires protected
  infrastructure to feel safe.

No Way No How: Non-cyclists who find cycling
  too dangerous regardless of infrastructure.
\end{verbatim}

\end{document}